\documentclass[runningheads]{llncs}
\usepackage[T1]{fontenc}
\usepackage{graphicx}
\usepackage{booktabs}
\usepackage[misc]{ifsym}
\usepackage{xurl}  
\newcommand{\corr}{(\Letter)}
\usepackage{mwe}

\usepackage{multirow}
\usepackage{hyperref}
\usepackage[table]{xcolor}

\begin{document}

\title{Large Language Models in Legislative Content Analysis: A Dataset from the Polish Parliament}

\titlerunning{Large Language Models in Legislative Content Analysis}

\author{Arkadiusz Bryłkowski\inst{1} \and Jakub Klikowski\inst{2}\orcidID{0000-0002-3825-5514} \corr}

\authorrunning{A. Bryłkowski and J. Klikowski}

\institute{\email{arkadiusz.brylkowski@gmail.com}
\and
Department~of~Systems~and~Computer~Networks,
Wrocław~University~of~Science~and~Technology,
Wrocław, Poland
\email{jakub.klikowski@pwr.edu.pl}}

\maketitle

\begin{abstract}
Large language models (LLMs) are among the best methods for processing natural language, partly due to their versatility. At the same time, domain-specific LLMs are more practical in real-life applications. This work introduces a novel natural language dataset created by acquired data from official legislative authorities' websites. The study focuses on formulating three natural language processing (NLP) tasks to evaluate the effectiveness of LLMs on legislative content analysis within the context of the Polish legal system. Key findings highlight the potential of LLMs in automating and enhancing legislative content analysis while emphasizing specific challenges, such as understanding legal context. The research contributes to the advancement of NLP in the legal field, particularly in the Polish language. It has been demonstrated that even commonly accessible data can be practically utilized for legislative content analysis.

\end{abstract}

\section{Introduction}

In the era of digitalization, there is a growing interest in advanced technological tools that support decision-making processes, data analysis, and task automation. Large language models (LLMs) occupy a particularly significant position among these. Their ability to analyze and generate texts opens new possibilities in the field of legislative content analysis, especially in the context of interpretation, classification, and the examination of complex legal documents~\cite{Law_and_order}.

Modern legal systems, characterized by high complexity, are evolving dynamically. As a result, traditional analysis methods based on manual expert work do not always meet the demands of speed and objectivity. LLMs, leveraging advanced learning algorithms, can support this process~\cite{NLP_in_Legal}. In the context of legislative analysis in Poland, which follows the tradition of continental law, LLMs must address challenges specific to the local linguistic and legal landscape while maintaining high analytical accuracy~\cite{Prawo_polskie}. Despite their potential, LLMs face difficulties related to the ambiguity of legal language, specialized terminology, and complex legal structures. It is also essential to consider historical, cultural, and social contexts~\cite{wstep_NLP_in_legal}.

Implementing natural language processing (NLP) in the legal domain raises concerns about potential misclassification and misinterpretation errors and the risk of results being used contrary to the public interest. However, researchers emphasize the significance of LLM studies for science and society, highlighting the necessity of a responsible approach to developing and deploying these technologies~\cite{wstep_NLP_in_legal2}. The primary objective of this study was to create a dataset for analyzing legislative content in the Polish legal system using LLMs. The research focused on constructing this dataset and subsequently analyzing it. Three NLP tasks were formulated by acquiring data from the official websites of legislative bodies to evaluate the effectiveness of selected models in the legal context.

The main contribution of this work:

\begin{enumerate}
    \item Legislative Data Acquisition -- Datasets were created using the Sejm RP API~\cite{Sejm_api} and publicly available information from the websites of the Polish Sejm and Senate~\cite{senat}.
    \item Design three benchmark datasets with different tasks to evaluate the natural language processing performance of language models in the legislative content domain for Polish:
    \begin{itemize}
        \item PPC (Predict Paper Category) -- multi-label classification of documents based on their thematic categories.
        \item PPO (Predict Paper Outcome) -- prediction of whether a legislative proposal requires amendments from the Senate (binary classification).
        \item STP (Summarize The Paper) -- automatic summarization of legislative drafts.
    \end{itemize}

    \item Preparation of a repository -- which collected scripts that allow downloading and updating data, data collected from 08.11.2011 to 01.03.2025, and scripts that enable the analysis of language models in solving the prepared tasks.
    
    \item LLM Evaluation -- An assessment of language models (including Polish variants of BERT and GPT) was conducted to measure their effectiveness in performing the tasks mentioned earlier. For this purpose, appropriate metrics were selected for the classification tasks and the summarization quality. In addition, statistical significance tests were performed on the obtained results.
\end{enumerate}

\section{Related works}

In the legal context, NLP faces numerous challenges related to the complexity of legal language. Key applications include predicting court case outcomes~\cite{LJP}, answering legal inquiries~\cite{LQA}, summarizing documents~\cite{ESOLT}, as well as searching for and comparing similar cases. One study discusses two approaches: embedding-based methods, which transform legal texts into numerical vectors to facilitate semantic analysis, and symbol-based methods, which utilize predefined symbols to represent legal concepts, enabling structured reasoning~\cite{NLP_in_Legal}.

In the context of this study, it is worth mentioning the authors of the EURLEX57K dataset, which contains 57,000 legislative documents from the EUR-LEX portal, annotated with approximately 4,300 EUROVOC labels~\cite{LMTC}. This dataset is used for large-scale multi-label text classification (LMTC) in the legal domain. The authors demonstrated that analyzing specific sections of documents can be sufficient for classification purposes, allowing for the efficient use of models such as BERT~\cite{BERT}. In the following years, an expansion of this dataset Multi-EURLEX~\cite{chalkidis-etal-2021-multieurlex} was created, where the new version covers many languages, including Polish. However, as the authors explain, these documents are based on official translations and were not initially written in multiple languages, which may cause some deviations from the native language. 

Building upon the EURLEX57K, authors introduced LEGAL-BERT, a family of BERT models tailored for the legal domain~\cite{LEGAL-BERT}. These models outperform the original BERT in tasks related to legal texts, emphasizing the importance of adapting language models to specific domains. The creators of LEGAL-BERT also developed the ECHR (European Court of Human Rights) dataset~\cite{ECHR}, an extended version of the work "Predicting judicial decisions of the European Court of Human Rights: a Natural Language Processing perspective"~\cite{ECHR1}. Ilias Chalkidis et al. published a dataset for predicting legal rulings of the European Court of Human Rights, comprising 11,500 cases, compared to Nikolaos Aletras et al., whose dataset contained only 6,500 cases. Researchers in China made similar efforts, developing the CAIL2018 dataset~\cite{CAIL2018}. CAIL2018 (Chinese AI and Law dataset 2018) is a significant legal dataset created for court ruling prediction tasks within the Chinese legal system. It includes over 2.6 million court cases based on actual court data from China.

Another example highlighting the significant role of dataset collection is the GLUE benchmark~\cite{GLUE}. It comprises a diverse set of NLP tasks designed to assess models' natural language understanding capabilities across various aspects, including sentiment analysis, question answering, and linguistic inference. The authors of the LexGLUE~\cite{chalkidis-etal-2022-lexglue} dataset had a similar goal: assembling various datasets with NLP tasks to evaluate language models in the legal domain. Inspired by GLUE, Polish researchers developed the KLEJ (direct Polish translation from GLUE) benchmark, which contains nine tasks tailored for the Polish language, such as named entity recognition, sentiment analysis, and cyberbullying detection~\cite{KLEJ}. As part of KLEJ, the HerBERT model was introduced~\cite{HerBERT}, trained explicitly for Polish language processing. The LEPISZCZE project is another benchmark for the Polish language, designed to facilitate NLP model comparisons by providing a broad range of tasks and a modern API~\cite{LEPISZCZE}. 

The Parliamentary Discourse Corpus is another noteworthy project that publishes an extensive collection of Polish parliamentary transcripts from 1918 to the present, processed and annotated using linguistic tools. This project enables a comprehensive analysis of parliamentary language~\cite{KDP_corpus_paper3}. In evaluating large language models, MT-Bench and Chatbot Arena propose new approaches for testing and comparing models through complex questions and a crowdsourcing-based platform~\cite{MT-Bench}. The Polish Information Retrieval Benchmark (PIRB) is a comprehensive evaluation tool for Polish information retrieval methods, covering 41 tasks across various domains, including law~\cite{dadas2024pirb}. A more recent study, "Legal Knowledge Representation Learning"~\cite{LKRL}, focuses on representing legal knowledge, particularly in machine learning applications. The authors first introduce three typical tasks requiring extensive legal knowledge—judicial decision prediction, legal information retrieval, and legal question answering—before proposing example approaches to address these challenges. 

The presented studies and initiatives highlight the dynamic development of NLP tools and methods in the legal context and for the Polish language. Using specialized datasets, domain-specific models, and benchmarks enables more effective analysis and processing of complex legal texts, which is crucial for further research and practical applications.

\section{Datasets}

This section will present original data sets, which constitute the primary original contribution of this study. The developed datasets result from a comprehensive review and analysis of information available on government websites related to the legislative process. These data were carefully selected, processed, and compiled to facilitate the execution of the research tasks.

Following a review of publicly available data on Polish government websites, the official pages of the Sejm, Senate, and the Sejm API were chosen as primary sources~\cite{Sejm_api}. Data from the Dziennik Ustaw (Journal of Laws) were retrieved using an explicitly developed custom Python script. The collected data enabled the development of three unique tasks. Table~\ref{tab:datastat} shows an overview of proposed dataset statistics.

\begin{table}[]
\centering
\caption{Number of documents, mean sentences per document, and unique words per document for PPC, PPO, STP datasets}
\begin{tabular}{rccc}
                             & \textbf{PPC}             & \textbf{PPO}             & \textbf{STP}                   \\ \hline
\emph{Number of documents} & 3738        \cellcolor{gray!10} & 1533         & 1327       \cellcolor{gray!10}  \\ \hline
\emph{Mean sentences}      & 435         \cellcolor{gray!10} & 2049         &2123/142   \cellcolor{gray!10}\\ \hline
\emph{Unique words}        & 1882        \cellcolor{gray!10} & 4942         &4924/974   \cellcolor{gray!10}
\end{tabular}
\label{tab:datastat}
\end{table}

\subsection{PPC}

The first developed task is \textbf{PPC (Predict Paper Category)}, which focuses on the multi-label classification of publications issued by the Sejm. A key aspect of interest in this task is the keywords feature, which provides information about the document's topic, i.e., its label. 

\begin{figure}[t]
  \includegraphics[width=\columnwidth]{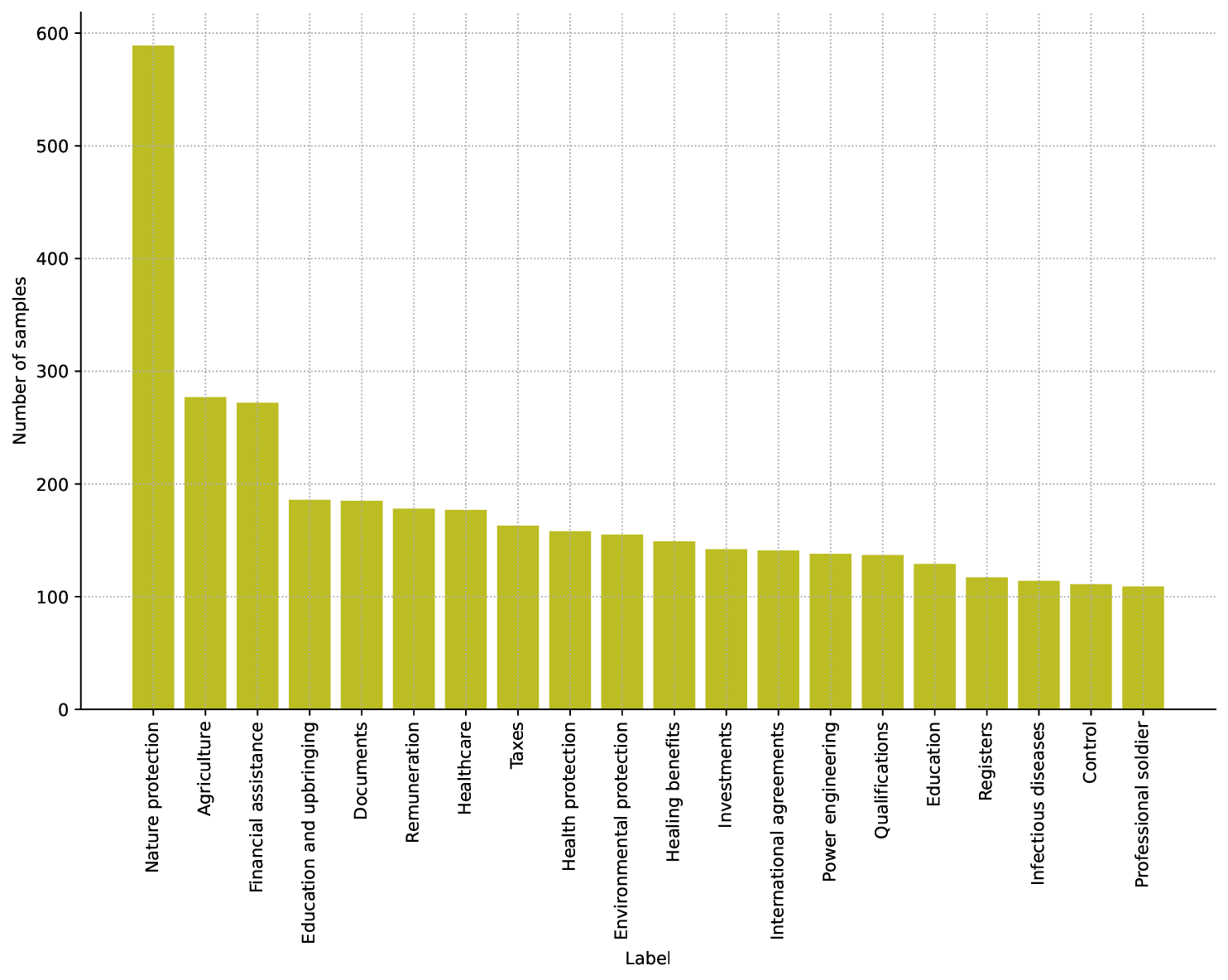}
  \caption{Histogram of labels frequency in the PPC dataset.}
  \label{fig:ppc_labels}
\end{figure}

\begin{figure*}[t]
  \includegraphics[width=1\linewidth]{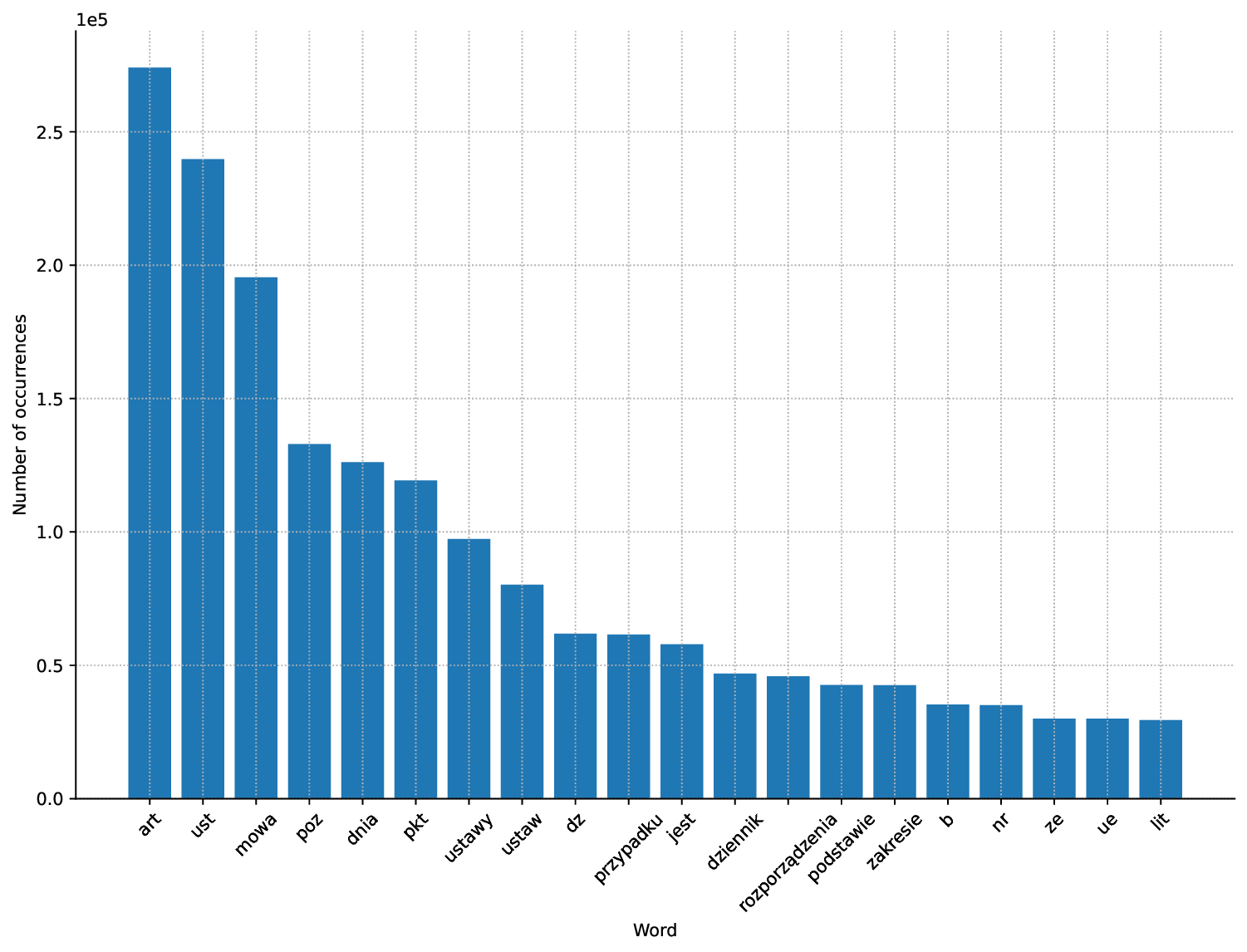} 
  \caption {Word frequency in legislative acts for the PPC task.}
  \label{fig:ppc}
\end{figure*}

An analysis of the PPC dataset, which comprises 3,738 documents with diverse labels, reveals its imbalanced nature. The dataset includes labels that may be considered synonyms, such as environmental protection and nature conservation. The total number of unique labels is 312, posing a challenge for the classification process. It was decided that only the top 20 labels would be used. Figure~\ref{fig:ppc_labels} shows the 20 most frequently occurring labels. Statistical analysis established that the average number of sentences per document in this dataset is 435, while the average number of unique words is 1,882. Figure~\ref{fig:ppc} illustrates the 20 most frequently occurring words.

\subsection{PPO}

The second developed dataset is \textbf{PPO (Predict Paper Outcome)}. This task represents a simplified approach to analyzing errors in legislative drafts. It involves binary classification and predicting whether a given legislative proposal requires amendments. The dataset comprises legislative drafts submitted by the Sejm and the Senate’s decision on whether the proposed bill requires modifications. This process takes advantage of the fact that when the Senate introduces amendments to bills, these modifications are also labeled accordingly.

\begin{figure*}[t]
  \includegraphics[width=1\linewidth]{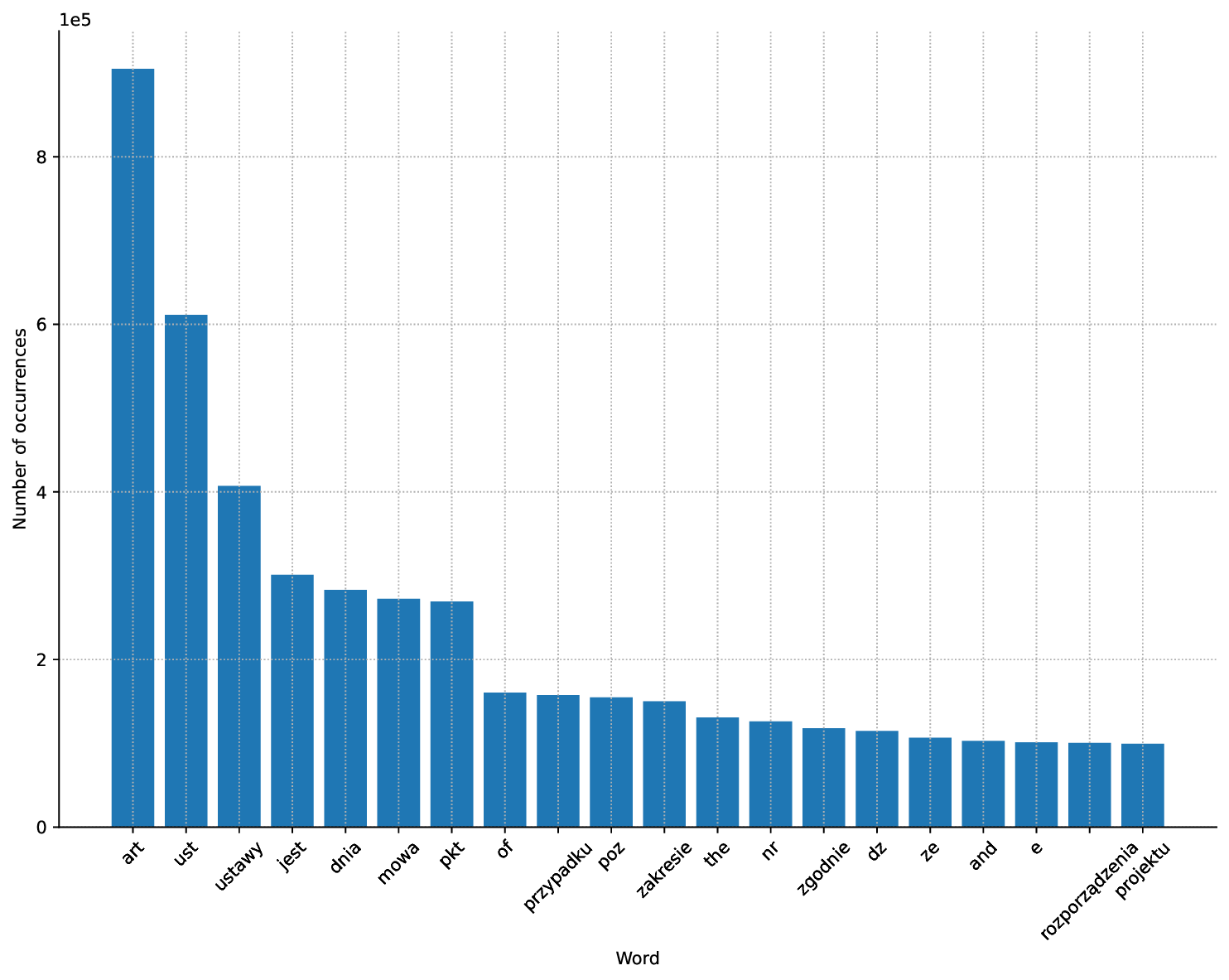} 
  \caption {Word frequency in legislative acts for the PPO task.}
  \label{fig:ppo}
\end{figure*}

As a result, the dataset includes 1,533 legislative proposals that did not require amendments and 839 proposals that the Senate modified. It is important to emphasize that the Senate’s introduction of amendments does not necessarily imply that they were ultimately adopted. 

Statistical analysis of this dataset established that the average number of sentences per document is 2,049, while the average number of unique words is 4,942. The figure~\ref{fig:ppo} illustrates the 20 most frequently occurring words in this dataset. 

\subsection{STP}

The final proposed dataset is \textbf{STP (Summarize The Paper)}. This task represents a classical natural language processing problem focused on summarizing a block of text. The dataset comprises 1,327 legislative proposals and an equal number of corresponding summary texts. The task was constructed using the content of legislative proposals from the Sejm’s website as input and the corresponding analyses published by the Bureau of Research (BAS) as summaries.

The purpose of the publications issued by BAS is to analyze legislative proposals initiated in the legislative process, provide members of parliament and Sejm bodies with the necessary information for fulfilling their duties, and raise awareness among lawmakers about key social and economic issues from a legislative perspective. These documents have been used as summaries of legislative proposals.

Such analyses typically consist of multiple sections, including key points, benefit assessments, and conclusions. Two approaches can be applied to this problem: the first involves using the entire BAS document as a summary (including all paragraphs). In contrast, the second focuses on extracting only the most relevant sections, such as key points or summary. In this study, the decision was made to use the entire summary, as a review of several cases revealed that not all BAS analyses contained these specific sections, and their format frequently varied over the years.

\begin{figure*}[t]
\centering
  \includegraphics[width=1\linewidth]{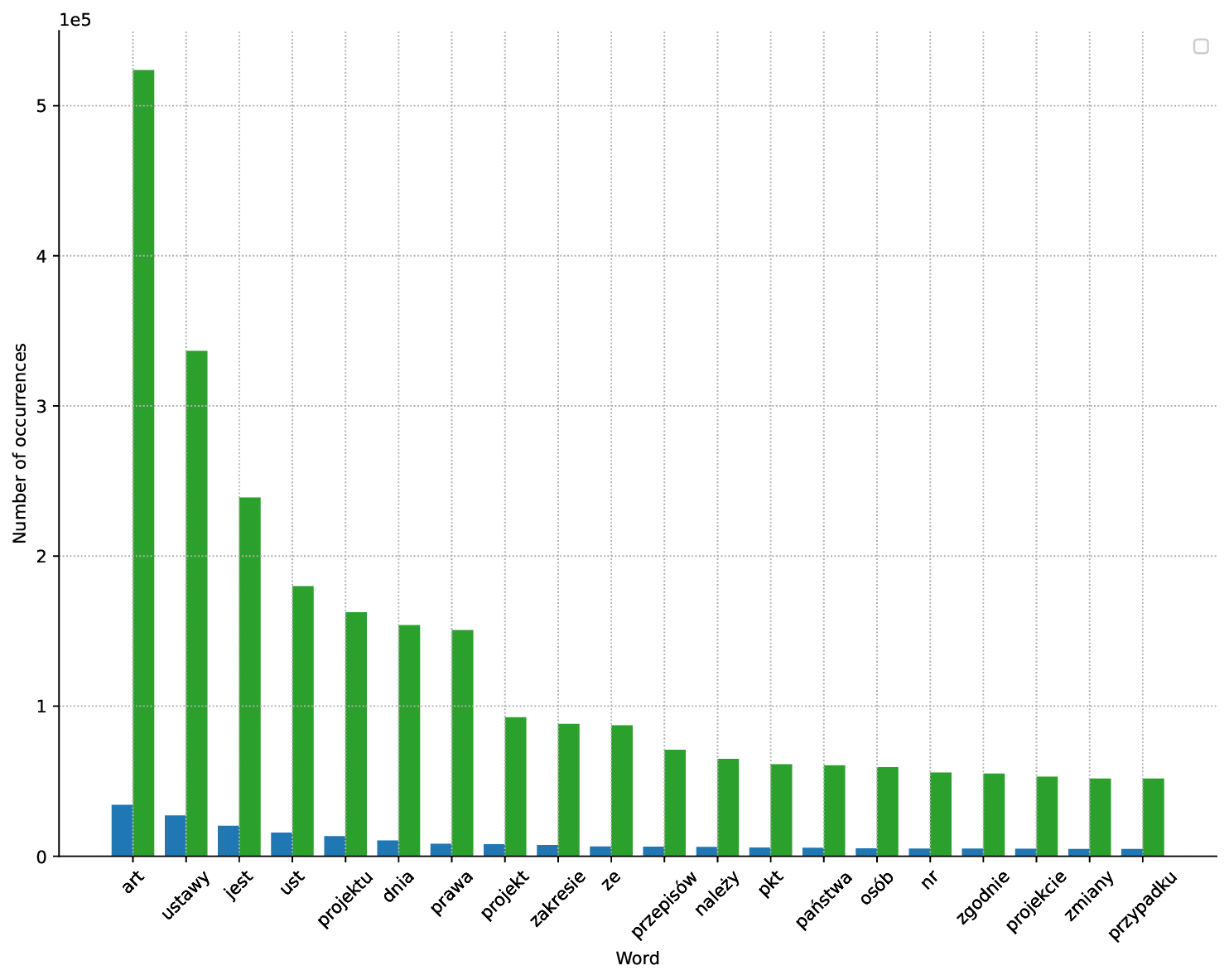} 
  \caption {Word frequency in legislative acts for the STP task.}
  \label{fig:stp}
\end{figure*}

An analysis of this dataset established that the input texts contain an average of 2,123 sentences and 4,924 unique words. In contrast, the summaries contain an average of 142 sentences and 974 unique words. The word frequency distribution shown in Figure (\ref{fig:stp}) illustrates the reduction in the number of words used. 

\section{Experimental evaluation}

This section presents a review of several baseline models that were evaluated using the proprietary datasets described in the previous chapters. The objective was to examine the extent to which the type of data used during the training of each model influences their effectiveness in the \textbf{PPC}, \textbf{PPO}, and \textbf{STP} tasks.

\subsection{Proposed method}
Experiments for all tasks were trained and evaluated on {\texttt{Kaggle}} using a dual 2XP100 GPU setup. All experiments were performed using five-fold cross-validation.

The models selected for the preliminary evaluation of the PPC and PPO tasks: 
\begin{itemize}
    \item \textsc{HerBERT}~\cite{mroczkowski-etal-2021-herbert} pre-trained on Polish corpora (\textsc{herbert-base-cased}\footnote{\href{https://huggingface.co/allegro/herbert-base-cased}{\url{https://huggingface.co/allegro/herbert-base-cased}}})
    \item \textsc{PL-RoBERTa}~\cite{dadas2020pretraining} pre-trained on Polish corpora (\textsc{polish-roberta-base-v2}\footnote{\href{https://huggingface.co/sdadas/polish-roberta-base-v2}{https://huggingface.co/sdadas/polish-roberta-base-v2}})
    \item \textsc{RoBERTa}~\cite{DBLP:journals/corr/abs-1907-11692} trained on the Wikipedia corpus in multiple languages and the BookCorpus
    \item \textsc{PL-GPT2}~\cite{polish-nlp-resources} pre-trained on Polish corpora (\textsc{sdadas/polish-gpt2-small}\footnote{\href{https://huggingface.co/sdadas/polish-gpt2-small}{https://huggingface.co/sdadas/polish-gpt2-small}})
    \item \textsc{GPT2}~\cite{radford2019language} pre-trained on English language data (\textsc{ComCom/gpt2-small}\footnote{\href{https://huggingface.co/ComCom/gpt2-small}{https://huggingface.co/ComCom/gpt2-small}})
\end{itemize}

The models listed above were fine-tuned over 10 epochs to improve their classification ability for a given task. The maximum token sequence length was set to 512, and the AdamW optimizer was used with parameters: lr = 2e-5 and weight\_decay = 0.01. The evaluation metrics for the PPO and PPC tasks included:

\begin{itemize}
\item Accuracy – the proportion of correctly classified samples; a less informative metric in cases of class imbalance.
\item Recall – the model’s ability to detect all true positive cases.
\item Precision – indicates the proportion of model-classified “positive” examples that are genuinely positive.
\item F1 (micro) – a recommended metric for imbalanced data, considering all correct and incorrect predictions for each label.
\item ROC AUC (micro) – assesses model performance at various decision thresholds, providing a multi-dimensional view of classification effectiveness.
\end{itemize}

For the STP task, the evaluation focused on generative models that are better suited for text summarization tasks. The following models were used:

\begin{itemize}
\item \textsc{t5-small}~\cite{2020t5} implementation from HuggingFace repository\footnote{\href{https://huggingface.co/google-t5/t5-small}{\url{https://huggingface.co/google-t5/t5-small}}}
\item \textsc{PLBart}~\cite{PLBART} implementation from HuggingFace repository\footnote{\href{https://huggingface.co/docs/transformers/model_doc/plbart}{\url{https://huggingface.co/docs/transformers/model_doc/plbart}}}
\item \textsc{bart-large}~\cite{DBLP:journals/corr/abs-1910-13461} implementation from HuggingFace repository\footnote{\href{https://huggingface.co/facebook/bart-large}{\url{https://huggingface.co/facebook/bart-large}}}
\end{itemize}

To assess the quality of the generated paraphrases, the ROUGE metric was applied, measuring the degree of overlap between the generated text and the reference text, thereby indicating their similarity.
These experiments were designed to evaluate how much the training data (Polish vs. multilingual corpora) influence the performance of different models across various natural language processing tasks. Detailed results and their discussion are presented in the subsequent sections of this study.

\subsection{Research Questions}
To gain a deeper understanding of the characteristics of the developed dataset and the effectiveness of the applied methods, the following research questions were formulated:
\begin{enumerate}
    \item[RQ1:] What characteristics of the developed dataset contribute to its complexity in natural language processing tasks?

    \item[RQ2:] Do language models specifically designed for the Polish language have superior predictive performance compared to multilingual models in the tasks proposed in this study?  

    \item[RQ3:] Are large language models capable of automatically summarizing legislative documents in the Polish language within the proposed dataset?  
\end{enumerate}

\subsection{PPC}

The results obtained for the PPC task presented in Table~\ref{tab:PPC} show the mean values with the standard deviation for the selected metrics and the five language models tested. In addition, the advantages over other methods with statistical significance performed with paired Student's t-tests are presented above the results. 

\begin{table*}[h]

    \centering
    \caption{Mean and standard deviation for selected metrics on the PPC task from 5-fold cross-validation. Above results advantage of a certain method over others with statistical significance.}
    \resizebox*{\textwidth}{!}{
    \begin{tabular}{ccccccc}
        \# & \textbf{Model} & \textbf{Accuracy}  & \textbf{Recall}  & \textbf{Precision}  & \textbf{F1 Score}  & \textbf{ROC AUC} \\ \hline 
        \multirow{2}{*}{\textit{1}} & \multirow{2}{*}{\textsc{HerBERT}} & \textit{3,4} \cellcolor{gray!10} & \textit{3,4} & \textit{3,4,5} \cellcolor{gray!10} & \textit{3,4} & \textit{3,4} \cellcolor{gray!10} \\
         & & 0.759$\pm$0.022 \cellcolor{gray!10} & 0.849$\pm$0.021 & 0.878$\pm$0.005 \cellcolor{gray!10} & 0.863$\pm$0.010 & 0.921$\pm$0.010 \cellcolor{gray!10} \\ \hline 
        \multirow{2}{*}{\textit{2}} & \multirow{2}{*}{\textsc{PL-RoBERTa}} & \textit{3,4} \cellcolor{gray!10} & \textit{3,4} & \cellcolor{gray!10} & \textit{3,4} & \textit{3,4} \cellcolor{gray!10} \\
         & & 0.746$\pm$0.019 \cellcolor{gray!10} & 0.848$\pm$0.021 & 0.860$\pm$0.015 \cellcolor{gray!10} & 0.854$\pm$0.015 & 0.919$\pm$0.011 \cellcolor{gray!10} \\ \hline 
        \multirow{2}{*}{\textit{3}} & \multirow{2}{*}{\textsc{RoBERTa}} & \cellcolor{gray!10} & & \cellcolor{gray!10} & & \cellcolor{gray!10} \\
         & & 0.707$\pm$0.025 \cellcolor{gray!10} & 0.796$\pm$0.016 & 0.858$\pm$0.013 \cellcolor{gray!10} & 0.826$\pm$0.012 & 0.894$\pm$0.008 \cellcolor{gray!10} \\ \hline 
        \multirow{2}{*}{\textit{4}} & \multirow{2}{*}{\textsc{GPT2}} & \cellcolor{gray!10} & & \cellcolor{gray!10} & & \cellcolor{gray!10} \\
         & & 0.711$\pm$0.023 \cellcolor{gray!10} & 0.790$\pm$0.023 & 0.863$\pm$0.009 \cellcolor{gray!10} & 0.825$\pm$0.012 & 0.891$\pm$0.011 \cellcolor{gray!10} \\ \hline 
        \multirow{2}{*}{\textit{5}} & \multirow{2}{*}{\textsc{PL-GPT2}} & \textit{4} \cellcolor{gray!10} & \textit{3,4} & \cellcolor{gray!10} & \textit{4} & \textit{3,4} \cellcolor{gray!10} \\
         & & 0.746$\pm$0.022 \cellcolor{gray!10} & 0.837$\pm$0.012 & 0.857$\pm$0.012 \cellcolor{gray!10} & 0.847$\pm$0.012 & 0.914$\pm$0.007 \cellcolor{gray!10} \\
    \end{tabular}
    }
    \label{tab:PPC}
\end{table*}

This experiment involved three pre-trained language models (HerBERT, PL-RoBERTa, and PL-GPT2) adapted to process Polish. At first glance, the advantage of these particular models over the others is noticeable. In many of the pairwise combinations, the advantage achieves statistical significance. There is also a noticeable difference between the results of the PL-RoBERTa and RoBERTa models and between PL-GPT2 and GPT2. This outcome strongly suggests that adapting the base model to specific language processing is crucial for better modeling the word space by the language model.

The best result for the PPC task was obtained using the HerBERT model. HerBERT attained an F1 score of 0.863, an ROC AUC of 0.921, and an Accuracy of 0.759. At the same time, the RoBERTa model performed the weakest. On average, the results for all models remain relatively high (the lowest result is 0.707 for the accuracy metric, and the highest is 0.759). Nonetheless, there is still some margin for improvement, meaning the task is not trivial.

\subsection{PPO}

Table~\ref{tab:PPO} also shows mean values with standard deviations and marked significant statistical advantage for selected metrics but for experiments in the PPO task. The results obtained look different. On average, the results are at a relatively low level of about 0.6 accuracy, which in the case of dichotomies means a quality close to that of a random classifier. This observation also means the task is somewhat challenging for the large language models tested. The best results among all models were obtained by the HerBERT model, which more than once achieved a statistically significant advantage. At the same time, it is worth noting that the RoBERTa model obtains the lowest score. In this experiment, the advantage of models adapted to Polish is no longer so noticeable, except for the HerBERT model.

\begin{table*}[h]

    \centering
    \caption{Mean and standard deviation for selected metrics on the PPO task from 5-fold cross-validation. Above results advantage of a certain method over others with statistical significance.}
    \resizebox*{\textwidth}{!}{
    \begin{tabular}{ccccccc}
        \# & \textbf{Model} & \textbf{Accuracy}  & \textbf{Recall}  & \textbf{Precision}  & \textbf{F1 Score}  & \textbf{ROC AUC} \\ \hline 
        \multirow{2}{*}{\textit{1}} & \multirow{2}{*}{\textsc{HerBERT}} & \textit{2,4,5} \cellcolor{gray!10} & \textit{5} & \textit{2,4,5} \cellcolor{gray!10} & \textit{2,4,5} & \textit{2,4,5} \cellcolor{gray!10} \\
         & & 0.625$\pm$0.025 \cellcolor{gray!10} & 0.542$\pm$0.037 & 0.681$\pm$0.008 \cellcolor{gray!10} & 0.579$\pm$0.014 & 0.723$\pm$0.007 \cellcolor{gray!10} \\ \hline 
        \multirow{2}{*}{\textit{2}} & \multirow{2}{*}{\textsc{PL-RoBERTa}} & \cellcolor{gray!10} & & \cellcolor{gray!10} & & \cellcolor{gray!10} \\
         & & 0.557$\pm$0.059 \cellcolor{gray!10} & 0.507$\pm$0.086 & 0.638$\pm$0.015 \cellcolor{gray!10} & 0.524$\pm$0.029 & 0.676$\pm$0.028 \cellcolor{gray!10} \\ \hline 
        \multirow{2}{*}{\textit{3}} & \multirow{2}{*}{\textsc{RoBERTa}} & \cellcolor{gray!10} & & \cellcolor{gray!10} & & \cellcolor{gray!10} \\
         & & 0.520$\pm$0.123 \cellcolor{gray!10} & 0.341$\pm$0.228 & 0.603$\pm$0.082 \cellcolor{gray!10} & 0.388$\pm$0.232 & 0.680$\pm$0.040 \cellcolor{gray!10} \\ \hline 
        \multirow{2}{*}{\textit{4}} & \multirow{2}{*}{\textsc{GPT2}} & \cellcolor{gray!10} & & \cellcolor{gray!10} & & \cellcolor{gray!10} \\
         & & 0.523$\pm$0.030 \cellcolor{gray!10} & 0.471$\pm$0.045 & 0.616$\pm$0.013 \cellcolor{gray!10} & 0.494$\pm$0.021 & 0.659$\pm$0.016 \cellcolor{gray!10} \\ \hline 
        \multirow{2}{*}{\textit{5}} & \multirow{2}{*}{\textsc{PL-GPT2}} & \cellcolor{gray!10} & & \cellcolor{gray!10} & & \cellcolor{gray!10} \\
         & & 0.532$\pm$0.055 \cellcolor{gray!10} & 0.452$\pm$0.024 & 0.616$\pm$0.024 \cellcolor{gray!10} & 0.488$\pm$0.028 & 0.663$\pm$0.028 \cellcolor{gray!10} \\
    \end{tabular}
    }
    \label{tab:PPO}
\end{table*}

\subsection{STP}

The evaluation results on the test dataset for the STP task are presented in Table~\ref{tab:wyniki-STP}. The ROUGE-1, ROUGE-2, ROUGE-L, and ROUGE-LSum metrics indicate how well the generated summaries align with the reference texts. The best-performing model was T5, suggesting that it is the most effective in generating summaries for legislative documents compared to the other tested models. However, even for this model, the metric values remain unsatisfactory. The PLBart and BART models achieved slightly lower scores, which may be attributed to various factors such as model architecture, training methodology, and the specificity of the training data.

\begin{table*}
  \centering
  \caption{Average cross-validation results of models on the STP task.}
  \begin{tabular}{ccccc}
    \textbf{Model} & \textbf{Rouge-1} & \textbf{Rouge-2} & \textbf{Rouge-L} & \textbf{Rouge-LS} \\
    \hline
\textsc{plbart}         & $0.023$ \cellcolor{gray!10}  & $0.003$           & $0.023$  \cellcolor{gray!10}  & $0.023$                    \\ 
\textsc{bart}           & $0.027$ \cellcolor{gray!10}  & $0.012$           & $0.025$  \cellcolor{gray!10}   & $0.027$                   \\ 
\textsc{t}$5$             & \textbf{0.036} \cellcolor{gray!10}   & \textbf{0.013}  & $\textbf{0.034}$  \cellcolor{gray!10}  & \textbf{0.034}     \\ \hline
  \end{tabular}
  \label{tab:wyniki-STP}
\end{table*}

\section{Lessons Learned}

The experimental evaluation allowed the study of how modern natural language processing methods performed on proposed datasets. The results led to important observations and crucial insights about the data. To summarize the experimental outcomes, let us try to answer the research questions formulated at the beginning of this section.\\

\noindent \textbf{\textit{What characteristics of the developed dataset contribute to its complexity in natural language processing tasks?}}\\

In the text categorization multi-classification task in the PPC dataset, models achieved very high scores across all metrics. These high results may stem from the nature of the task, where models identify specific vocabulary patterns corresponding to different categories -- for example, the word "doctor" may be unique to the "healthcare" category. In contrast, the models performed poorly in the binary classification task in the PPO dataset. These low scores may be due to the complexity of the task. To classify whether a legislative draft requires amendments, a model must understand specific words and the overall context, the intent behind proposed changes, their benefits and drawbacks, and even linguistic nuances in the Polish legislative language. Greater difficulty may also mean greater value in this data and considerable potential for training new models using this content.\\

\noindent \textbf{\textit{Do language models specifically designed for the Polish language have superior predictive performance compared to multilingual models in the tasks proposed in this study?}}\\

Models pre-trained on Polish corpora demonstrate significantly (Tables~\ref{tab:PPC} and~\ref{tab:PPO}) better performance in the designed Polish-language tasks (PPC and PPO). The HerBERT model outperformed most selected reference models in the PPO and PPC tasks. Notably, in the case of GPT2 models, pre-training on a Polish dataset appears to have positively impacted the performance of PL-GPT2, as evidenced by results in Table~\ref{tab:PPC}. PL-GPT2 achieved statistically significantly higher results in the PPC task than in the GPT model. Similar observations are visible for the RoBERTa and PL-RoBERTa models, where this advantage is statistically significant in favor of the model trained on Polish language content.\\

\noindent \textbf{\textit{Are large language models capable of automatically summarizing legislative documents in the Polish language within the proposed dataset?}}\\

The results for the STP task, evaluating text summarization models using ROUGE metrics, indicate that all tested models struggled to generate high-quality summaries. T5 achieved the best scores across all ROUGE metrics, suggesting a slight advantage over other models. BART performed comparably but with slightly lower scores than T5. One key factor contributing to T5's superior performance may be its training approach. T5 was trained on large-scale datasets in a manner that enhances coherence and fluency in text generation. Its architecture was optimized for diverse NLP tasks, which may provide an edge in processing and summarizing texts. While BART is also an advanced model, its architecture and training methodology differ. BART is an autoencoder-based model that combines bidirectional text processing with autoregressive text generation. Although adequate for reconstruction and transformation tasks, its architecture may not be as versatile or flexible as T5 in diverse NLP applications. The evaluation results may also reflect computational constraints (such as a limited context window) during experimentation.

\section{Conclusions}

This work proposes a new dataset consisting of three separate tasks covering the topic of natural language processing in the Polish legislative language domain. The data was extracted from publicly available laws and acts establishing Polish law, then organized and pre-processed to improve the quality of the content. An analysis using selected large language models was performed to determine the usefulness and the difficulty level of the proposed NLP tasks for the Polish language. The provided source code repository includes an implementation of the experimental environment and scripts to update this data to ensure the replicability of the research and the ability to develop this dataset. 

The key findings of this study highlight the high potential of large language models (LLMs) in automating legislative analysis and the challenges associated with the legal context, including the necessity for a deep understanding of legal language specificity. This work contributes to advancing NLP technologies in Polish law, emphasizing the practical use of publicly available data for legislative analysis.

Future plans include expanding the existing datasets. For the PPC dataset, merging synonymous categories could reduce the number of labels. The PPO dataset could be extended to include amendments introduced by the Sejm and the President’s decision on whether a bill is enacted. In contrast, the STP task requires standardization of the analyses prepared by the Bureau of Research of the Sejm, as the formatting of these summaries varies significantly across years. Additionally, collaboration with government legislative websites could enable the creation of further specialized NLP tasks.

\begin{credits}
\subsubsection{Ethics Considerations} This work makes available some data resources, and raising issues related to potential ethical implications is important. The most important information related to this collection is that it is based on data that is publicly available (to anyone interested) through the “Sejm RP API”~\cite{Sejm_api} service and the official website of the Polish Sejm\footnote{\href{https://www.sejm.gov.pl/sejm10.nsf/page.xsp/copyright}{\url{https://www.sejm.gov.pl/sejm10.nsf/page.xsp/copyright}}}. Therefore, we do not see any ethical concerns regarding using this data for analysis through natural language processing methods. According to our knowledge, these documents are subjected to in-depth analysis by groups of experts involved in legislation in the Polish government and are repeatedly read and approved by the Polish government before they are made public. This guarantees they are at a high linguistic level and free of potential errors.

\subsubsection{\ackname} This work was supported by the statutory funds of the Department of Systems and Computer Networks, Faculty of Information and Communication Technology, Wroclaw University of Science and Technology.

\end{credits}

\end{document}